%% file: acl_latex.tex
\pdfoutput=1

\documentclass[11pt]{article}

\usepackage[preprint]{acl}

\usepackage{times}
\usepackage{latexsym}

\usepackage[T1]{fontenc}

\usepackage[utf8]{inputenc}

\usepackage{microtype}

\usepackage{inconsolata}

\usepackage{graphicx}
\usepackage{natbib}
\usepackage{wrapfig,lipsum}
\usepackage{hyperref}
\usepackage{url}
\usepackage{graphicx}
\usepackage{booktabs}
\usepackage{bbding}
\usepackage{pifont}
\usepackage{multirow}
\usepackage{multicol}
\usepackage{amsmath}
\usepackage{amssymb}
\usepackage{verbatim}
\usepackage{fancyvrb}
\usepackage{pythonhighlight}
\usepackage{fvextra}
\usepackage{tcolorbox}
\tcbuselibrary{breakable}
\usepackage{longtable}
\usepackage{tabularray}
\usepackage{framed}
\usepackage{caption}
\usepackage{markdown}
\usepackage{xspace}
\usepackage{color}
\usepackage[table]{xcolor}
\usepackage{arydshln}
\tcbuselibrary{listings} 
\usepackage{enumitem}
\usepackage{stfloats}
\usepackage{makecell}

\newcommand{\method}{\textsc{UltraIF}\xspace}
\newcommand{\composer}{UltraComposer\xspace}

\title{\textsc{UltraIF}:  Advancing Instruction Following from the Wild}

\author{Kaikai An\textsuperscript{1,3}\thanks{~~Equal contribution}, Li Sheng\textsuperscript{2,3$*$}, Ganqu Cui\textsuperscript{3$\dag$},  Shuzheng Si\textsuperscript{2}, \textbf{Ning Ding\textsuperscript{2}},\\
\textbf{Yu Cheng\textsuperscript{3},}  \textbf{Baobao Chang\textsuperscript{1}\thanks{~~Corresponding author}} \\
\textsuperscript{1} Peking University \textsuperscript{2} Tsinghua University \textsuperscript{3} Shanghai AI Lab \\
\texttt{ankaikai@stu.pku.edu.cn, cuiganqu@pjlab.org.cn, chbb@pku.edu.cn}
}

\begin{document}
\maketitle

\begin{abstract}
Instruction-following made modern large language models (LLMs) helpful assistants. 
However, the key to taming LLMs on complex instructions remains mysterious, for that there are huge gaps between models trained by open-source community and those trained by leading companies. 
To bridge the gap, we propose a simple and scalable approach \textsc{UltraIF} for building LLMs that can follow complex instructions with open-source data. 
\textsc{UltraIF} first decomposes real-world user prompts into simpler queries, constraints, and corresponding evaluation questions for the constraints. Then, we train an \textit{\composer} to compose constraint-associated prompts with evaluation questions. 
This prompt composer allows us to synthesize complicated instructions as well as filter responses with evaluation questions. 
In our experiment, for the first time, we successfully align LLaMA-3.1-8B-Base to catch up with its instruct version on 5 instruction-following benchmarks without any benchmark information, using only 8B model as response generator and evaluator. The aligned model also achieved competitive scores on other benchmarks. Moreover, we also show that \textsc{UltraIF} could further improve LLaMA-3.1-8B-Instruct through self-alignment, motivating broader use cases for the method. Our code is available at \url{https://github.com/kkk-an/UltraIF}.
\end{abstract}

\input{sections/intro}
\input{sections/ultraif}

\input{sections/experiments}

\input{sections/relatedwork}

\section{Conclusion}
In this paper, we propose \method, a scalable and effective approach for synthesizing high-quality instruction-following data. By decomposing human instructions into simplified queries, constraints, and corresponding evaluation questions, we train \textit{\composer} that enables the efficient generation of constraint-aware instructions. Across two different settings, \method demonstrates strong performance across five instruction-following benchmarks and four general benchmarks. Extensive experiments conducted on LLaMA-3.1-8B-Instruct further highlight \method's potential for self-alignment. Most importantly, we are the first to optimize the LLaMA-3.1-8B-Base model to match the instruction-following capabilities of its Instruct counterpart, underscoring the effectiveness and potential of our approach.

\newpage
\section*{Limitations}
Due to limitations in time and computational resources, \method has not yet been evaluated on a wider range of backbone models or on models with larger parameter scales. Nevertheless, the current experimental results provide sufficient evidence of its generalizability across different foundational architectures. Additionally, since the full pipeline depends on LLMs for supervision, the data generation process may involve limited controllability and introduces potential risks related to consistency and reliability. To minimize these risks, we apply responses filtering to ensure the quality and stability of the generated data.

\section*{Acknowledgement}
We thank all reviewers for their great efforts. This work is supported by the National Science Foundation of China under Grant No.61876004.

\bibliography{custom}

\newpage
\appendix
\input{sections/appendix}

\end{document}

%% file: sections/intro.tex
\section{Introduction}

Large language models~\citep{grattafiori2024llama3herdmodels, openai2024gpt4technicalreport} have demonstrated remarkable capabilities, especially in following complex instructions. While modeling such ability is crucial, the technical details and the instruction datasets used in state-of-the-art LLMs remain mysterious. 
For example, LLaMA3 \citep{grattafiori2024llama3herdmodels} reportedly leverages instruction-following data at the tens of millions scale but has not been open-sourced. This lack of transparency has resulted in a significant gap between research community and leading companies.

\begin{figure}[!t]
    \centering
    \includegraphics[width=0.95\linewidth]{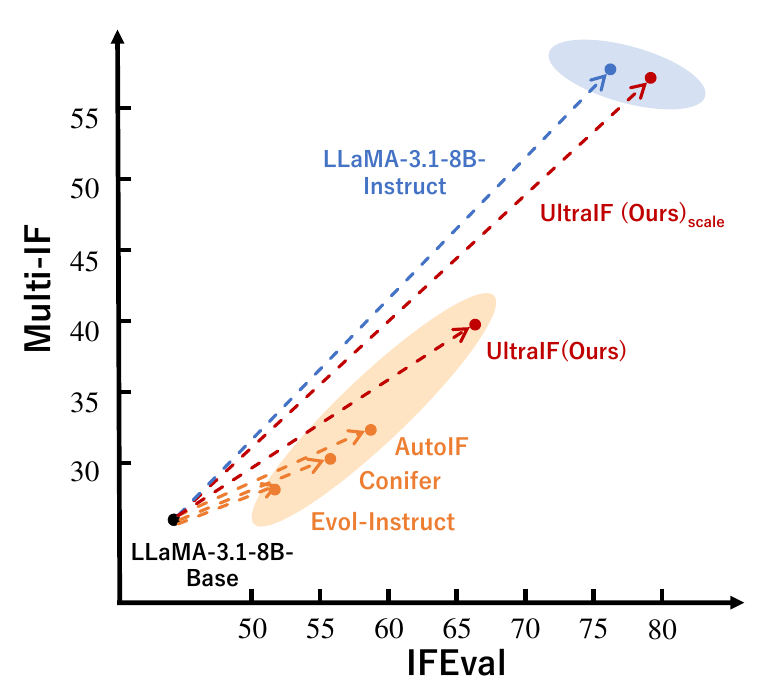}
    \caption{Instruction-following performance comparison of \method against baselines.}
    \label{fig:intro_performance}
    \vspace{-6mm}
\end{figure}

Recent efforts in aligning LLMs to follow instructions have focused on creating high-quality instruction-following data. 
On the one hand, \citet{wei2021finetuned, no_robots,jiang2023followbench} involve human annotators in developing instructions and manually crafting corresponding responses. While effective, these methods are label-intensive, heavily reliant on human expertise, and face challenges in scalability and cost efficiency.
On the other hand, \citet{xu2023wizardlm,wang-etal-2023-self-instruct,sun2024conifer,dong2024self} attempt to leverage LLMs to automatically construct high-quality instruction data.
Specifically, \citet{xu2023wizardlm,sun2024conifer} guide LLMs to generate constraints and evolve initial instructions into more complex forms.
However, these LLMs-driven methods heavily rely on models' instruction-evolving capability and overemphasize instruction complexity, ultimately hindering the diversity of evolved instructions and the correctness of generated responses.
To improve this, \citet{wang-etal-2023-self-instruct,dong2024self} introduce handcrafted constraints inspired by human priors to guide LLMs.
For instance, \citet{dong2024self} introduces constraints that can be verified by code execution to ensure response correctness.
However, these handcrafted constraints introduce rigidity, 
leading to homogeneous instructions and making it narrow in encompassing more complex or diverse instructions (e.g., \textit{write in Shakespeare's tone}).
As a result, scaling such instruction with correct responses remains a significant challenge, limiting the applicability of modeling the distribution of instructions from real-world users.

In this paper, we propose \method, a simple and scalable method which synthesizes high-quality instruction-following data.
The core idea of \method is to decompose real-world user instructions for both constraints and evaluation questions, then train a composer model to synthesize diverse and complex instructions with verification questions.
To achieve this, we first utilize LLM to decompose human instructions into simplified instructions and their associated constraints. 
For each constraint, the LLM further generates the corresponding evaluation question to verify whether the upcoming response meets the requirement.
With these components, we train \textbf{\textit{\composer}}, which takes the simplified instruction as input and outputs the original instruction along with its evaluation question.
In this way, the composer learns to evolve instructions with verifiable constraints, and benefits from the generalization ability of LLMs rather than handcrafted rules.
With the composer, \method could make any instruction more complicated to synthesize a large-scale and diverse dataset. The evaluation questions further help with quality control in rejection sampling and preference learning~\citep{rafailov2024direct,chen2024noise}.

Through comprehensive experiments, we demonstrate that \method significantly enhances the instruction-following capabilities of LLMs with high scalability and cost efficiency. Our evaluation, conducted on the LLaMA-3.1-8B model across five instruction-following datasets, confirms \method's strong alignment with general instructions. Notably, as shown in Figure \ref{fig:intro_performance}, by scaling up the training data, we achieve a milestone, optimizing the LLaMA-3.1-8B-Base model to match the instruction-following ability of its instruct version. Additionally, we assess the generability of \method by evaluating it on mathematical, reasoning, coding, and general conversation domains. Furthermore, we explore the potential of self-alignment in \method by further optimizing the LLaMA-3.1-8B-Instruct model, and achieved sustaintial improvement.

\noindent The main contributions of our paper include:
\vspace{-2mm}
\begin{itemize}[leftmargin=*]
    \setlength\itemsep{0mm}
    \item We introduce \method, a simple and scalable approach that leverages real-world user instructions to train a composer model, \textit{\composer}, enabling the synthesis of complex and diverse instructions with correct responses.
    \item Our experiments demonstrate the strong performance of \method in handling complex instructions, surpassing all baselines under the same data budget while retaining general capabilities in domains such as mathematics, coding, and conversational tasks.
    \item We reach a new milestone by optimizing the LLaMA-3.1-8B-Base model to match the instruction-following abilities of its Instruct counterpart with only 200k data, and showcase the self-alignment potential by further optimizing the LLaMA-3.1-8B-Instruct model on it own.

\end{itemize}

%% file: sections/ultraif.tex
\begin{figure*}[t]
    \centering
    \includegraphics[width=\textwidth]{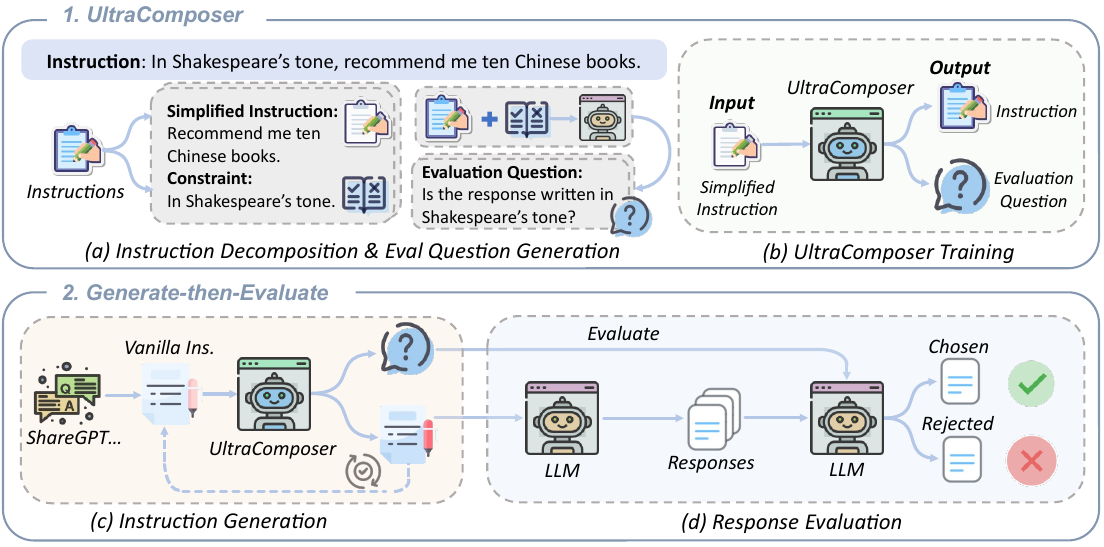}
    \caption{The framework of \method. Specifically, \method begins by training the \textbf{\textit{\composer}}, which decomposes real-world user instructions and evaluation questions. For (a), the given instruction can be decomposed into several pairs, such as the numeric constraint `ten books' and content constraint `Chinese books'. Next, \method adopts a \textbf{\textit{Generate-then-Evaluate}} process, where the composer iteratively adds multiple constraints to each collected instruction and then applies the evaluation questions for rejection sampling.
    }
    \label{fig:ultraif}
\end{figure*}

\section{\textsc{UltraIF}}
\subsection{Overview}

\method synthesizes high-quality instruction-following datasets in two stages. As shown in Figure \ref{fig:ultraif}, \method first constructs the \textbf{\textit{\composer}} by decomposing user instructions into simplified ones and constraints, along with corresponding evaluation questions (\S\ref{sec:composer}). This specialized composer facilitates the synthesis of instructions with more complex and diverse constraints, while the evaluation questions ensure the correctness and reliability of the generated responses.
Then, the \textbf{\textit{Generate-then-Evaluate}} process (\S\ref{sec:framework}) uses \textit{\composer} to incorporate constraints into instructions and assesses the generated responses using corresponding evaluation questions covering various quality levels.

\subsection{UltraComposer}
\label{sec:composer}
Previous studies \citep{xu2023wizardlm,sun2024conifer} that rely solely on LLMs are limited by the models' instruction-evolving ability, which restricts the diversity of synthetic instructions and compromises response accuracy. While \citet{wang-etal-2023-self-instruct,dong2024self} address response correctness through handcrafted constraints, this approach further limits instruction diversity.
In contrast, \method focuses on generating diverse, complex instructions with correct responses. To achieve this, we propose the \textit{\composer}, a specialized model to synthesize diverse instructions and generate corresponding evaluation questions. Building this composer model involves three key steps: instruction decomposition, evaluation question generation, and \textit{\composer} training.

\paragraph{Instruction Decomposition}
The decomposition process leverages LLMs to decompose complex instructions into different components.
These components consist of a set of simplified instructions paired with constraints that represent the underlying requirements of the original instruction. For example, as shown in Figure \ref{fig:ultraif} (a), the instruction ($X$) \textit{``In Shakespeare's tone, recommend me ten Chinese books.''} can be decomposed into the simplified instruction ($x_1$) \textit{``Recommend me ten Chinese books.''} and the paired constraint ($c_1$) \textit{``In Shakespeare's tone.''}, etc. This step is essential for disentangling intricate objectives into more structured elements, extending beyond basic format or content constraints \citep{dong2024self,wang-etal-2023-self-instruct}, and forming a foundation to model the distribution of real-world user instructions effectively.

\paragraph{Evaluation Question Generation}
While \citet{xu2023wizardlm,wang-etal-2023-self-instruct} focus on improving the complexity of instructions, omitting the quality of generated responses often leads to low-quality samples. Inspired by \citet{qin2024infobench}, we utilize LLM to generate evaluation questions for each constraint. Given the example, the evaluation question ($q_1$) would be \textit{``Is the response written in Shakespeare's tone?''}. These questions are designed to assess the generated responses for adherence to the constraints. 
This mechanism not only addresses the limitation of checking only programmatically verifiable constraints \citep{dong2024self}, but also improves the reliability of the response and its alignment with the original instruction.
\begin{equation}
    X \to \{(x_1, c_1, q_1),\ ...\ , (x_n, c_i,q_i)\},\ \ i\in \mathbb{N} 
\end{equation}

\paragraph{\composer Training}
With decomposed instructions and evaluation questions, we train \textit{\composer} to take a simplified query  ($x_i$) as input and generate the original instruction ($X$) with its evaluation question ($q_i$), denoted as Eq.\ref{eq:2}, as shown in Figure \ref{fig:ultraif} (b).
This enables \textit{\composer} to complicate instructions in a single step. Additionally, it enhances constraint diversity by incorporating not only the LLM's inherent knowledge but also distributions observed in real-world scenarios.
\begin{equation}
    \label{eq:2}
    UltraComposer(x_i) \to (X,\ q_i),\ \ i\in \mathbb{N}
\end{equation}

\subsection{Generate-then-Evaluate}
\label{sec:framework}
With \textit{\composer}, \method efficiently produces high-quality instruction-following data through a Generate-then-Evaluate process, encompassing both instruction generation and response evaluation to support both Supervised Fine-tuning and Preference Learning strategies.

\paragraph{Instrucion Generation}
\textit{\composer} adapts the augmentation process fully automated and aligns with human preferences. This step starts by collecting user instructions from existing datasets \citep{vicuna2023,OpenHermes,no_robots}, and then use the Composer to augment these instructions. As shown in Eq.\ref{eq:3}, this process can be conducted iteratively, enabling the generation of more complex and realistic instructions ($\bar{x}$) with multiple constraints, paired with corresponding evaluation questions ($\bar{q}$).
\begin{equation}
\label{eq:3}
\begin{aligned}
    \small
    UltraComposer(x^{(n)}) \to (\bar{x}^{(n)},\ \bar{q}^{(n)}),\ \ n \in \mathbb{N} \\
    x^{(n+1)} = \bar{x}^{(n)}, \quad \bar{q}^{(n+1)} = \bar{q}^{(n+1)} \cup \bar{q}^{(n)}
\end{aligned}
\end{equation}

\paragraph{Response Evaluation}
Next, we prompt LLMs to generate $K$ responses for each augmented instruction. As `LLM-as-judge' paradigm is prevalent \cite{zheng2023judging}, human can be replaced by LLMs to assess the quality of response, so the quality of generated responses is assessed by evaluation questions. This results in a dataset $\mathcal{D}_{data}$ comprising  $(\bar{x},\ \bar{q}, y_{chosen}, y_{rejected})$.
Ideally, this process requires only three to four calls to LLMs, significantly reducing the computational cost, achieves greater efficiency and incurs minimal costs when constraining large-scale datasets compared to previous research \citep{xu2023wizardlm, dong2024self}.

%% file: sections/experiments.tex
\section{Experiments}
\subsection{Experimental Setup}

\paragraph{Datasets and Baselines}
To train \textit{\composer}, we decompose instructions from ShareGPT \citep{vicuna2023} and generate corresponding evaluation questions by LLaMA-3.1-70B-Instruct.
In our experiments, we collect human instructions from existing open-source datasets, including ShareGPT, OpenHermes2.5, and No Robots \citep{OpenHermes,no_robots,vicuna2023}, and employ \textit{\composer} to complicate instructions and then generate responses.
For baselines, we reimplement existing methods using either public datasets \citep{sun2024conifer,xu2023wizardlm} or available implementations \citep{dong2024self}, and include a series of currently open and closed-source LLMs. More details are in Appendix \ref{sec:dataset}.

\paragraph{Evaluation}
We evaluate \method on five instruction-following benchmarks, including IFEval \citep{zhou2023instruction}, Multi-IF \citep{he2024multi}, InfoBench \citep{qin2024infobench}, FollowBench \citep{jiang2023followbench}, and LiveBench \citep{white2024livebench}. While IFEval and Multi-IF focus on testing verifiable instructions using functions, the others extend to more general instructions that need to be evaluated by LLMs. Additionally, we further test the general ability of \method such as mathematical \citep{chen2021evaluating}, reasoning \citep{suzgun2022challenging}, coding \citep{cobbe2021training}, and general interaction capabilities \citep{li2024crowdsourced}.
The details about benchmarks are provided in Appendix \ref{sec:benchmark}.

\paragraph{Experimental Settings}
We fine-tune LLaMA-3.1-8B-Instruct to build our \textit{UltraComposer}.
Subsequently, we explore two settings to implement our training strategies listed in Appendix \ref{sec:strategy}, including Supervised Fine-tuning and Preference Learning. The implementation details are listed in Appendix \ref{sec:train_details}. And the prompts about instruction decomposition and evaluation question generation are provided in Appendix \ref{sec:prompt}.
\vspace{-2mm}
\begin{itemize}[leftmargin=*]
    \setlength\itemsep{-1mm}
    \item \textbf{Strong-to-Weak.} In this setting, knowledge is distilled from a larger model to a smaller one. For \method and baselines, we leverage LLaMA-3.1-70B-Instruct for response generation and evaluation and then train LLaMA-3.1-8B-Base.
    \item \textbf{Self-Alignment.} We replace the supervision model with Llama-3.1-8B-Instruct.
\end{itemize}

\begin{table*}[!t]
    \centering
    \resizebox{\textwidth}{!}
    {
    \begin{tabular}{lc*{7}{>{\centering\arraybackslash}p{1cm}}*{3}{>{\centering\arraybackslash}p{1.7cm}}}
    \toprule
    \multirow{2}{*}{\textbf{Method}} & \multirow{2}{*}{\textbf{\#Data}} &  \multicolumn{4}{c}{\textbf{IFEval}} & \multicolumn{3}{c}{\textbf{Multi-IF}} & \textbf{InfoBench} & \textbf{LiveBench} & \textbf{FollowBench} \\ 
    \cmidrule(r){3-6}  \cmidrule(r){7-9} \cmidrule(r){10-10}  \cmidrule(r){11-11} \cmidrule(r){12-12}   
    & & Pr(S) & Pr(L) & Ins(S) & Ins(L) & Turn1 & Turn2 & Turn3 & DRFR & Score & SSR \\ 
    \midrule
    GPT-4$^\dag$ & - & 76.90  & 79.30  & 83.60  & 85.40  & 81.50  & 70.50  & 60.90  & 89.40  & 69.40 &  78.60 \\ 
    \rowcolor{green!5}  LLaMA-3.1-8B-Instruct$^\dag$ & - & 69.13  & 74.86  & 77.46  & 81.65  & 68.54  & 59.63  & 51.26  & 81.33  & 57.10 & 63.41 \\
    \midrule
    \rowcolor{pink!13}  \multicolumn{12}{c}{\textbf{\textit{Strong-to-Weak}    (Supervisor: GPT-4o-mini)}} \\ 
    \midrule
    SPaR \citeyearpar{cheng2024spar}$^\ddag$ & 8k & 54.71  & 58.59  & 64.86  & 68.70  & 55.37  & 36.22  & 27.23  & 78.61  & 50.80 & 59.37 \\ 
    \midrule
    \rowcolor{pink!13}  \multicolumn{12}{c}{\textbf{\textit{Strong-to-Weak}    (Supervisor: LLaMA-3.1-70B-Instruct)}} \\ 
    \midrule
    LLaMA-3.1-8B (ShareGPT) & 10k & 43.99  & 54.34  & 54.32  & 64.39  & 44.69  & 25.11  & 18.50  & 81.56  & 33.20 & 59.59 \\ 
    Evol-Instruct \citeyearpar{xu2023wizardlm}$^\ddag$ & 10k & 41.96 & 45.66 & 54.44 & 58.03 & 39.03 & 24.34 & 19.14 & 75.74 & 44.90 & 43.87 \\ 
    Conifer \citeyearpar{sun2024conifer}$^\ddag$ & 13k & 46.40 & 51.02 & 58.51 & 62.59 & 44.91 & 25.83 & 17.95 & 75.73 & 45.60 & 52.42 \\
    \textsc{AutoIF} \citeyearpar{dong2024self}$^\ddag$ & 10k & 47.13  & 56.93  & 57.55  & 67.02  & 47.63  & 27.53  & 20.53  & 80.62  & 40.50  & \textbf{60.41}  \\ 
    \rowcolor{gray!13} \method & ~ & ~ & ~ & ~ & ~ & ~ & ~ & ~ & ~ & ~ & ~\\
     + SFT & 10k & 53.97  & 58.59  & 64.15  & 68.82  & 52.55  & 29.34  & 22.29  & 81.91  & 42.20  & 59.50  \\ 
     \rowcolor{blue!5} \hspace{6pt} + Iterative DPO & 8k & \textbf{58.22}  & \textbf{65.25}  & \textbf{68.11} & \textbf{74.22}  & \textbf{58.14}  & \textbf{35.65}  & \textbf{26.55} & \textbf{83.56}  & \textbf{49.50} & 59.99   \\ 
    \midrule
    \rowcolor{pink!13} \multicolumn{12}{c}{\textbf{\textit{Self-Alignment}    (Supervisor: LLaMA-3.1-8B-Instruct)}} \\ 
    \midrule
    \rowcolor{gray!13} \method & ~ & ~ & ~ & ~ & ~ & ~ & ~ & ~ & ~ & ~ & ~\\ 
     + SFT & 10k & 55.82 & 58.78 & 66.18  & 69.54 & 55.59 & 36.72 & 28.07 & 77.78 & 46.60 & 55.88\\ 
     \hspace{6pt} + Iterative DPO & 8k & 56.93 & 64.14 & 66.66 & 73.02 & 58.63 & 42.04 & 31.20 & 79.86 & 54.20 & 58.56\\
     \hdashline[2pt/3pt]
     + SFT $scale\ up$ & 181k & 69.87  & 72.46  & 77.46  & 80.22  & 66.24  & 53.66  & 42.19  & 79.20 & 51.40  & 59.93 \\ 
     \rowcolor{blue!5} \hspace{6pt} + Iterative DPO & 20k & \textbf{71.35}  & \textbf{75.42}  & \textbf{79.38}  & \textbf{83.09} & \textbf{69.63} & \textbf{58.28} & \textbf{46.86} & \textbf{80.70} & \textbf{56.00} & \textbf{62.55 } \\ 
    \bottomrule
    \end{tabular}
    }
    \caption{The main results on five instruction-following benchmarks. 
    Results marked with $\dag$ are sourced from the original benchmarks, and $\ddag$ represents we reimplement the methods.}
    \label{tab:main_instruction}
    \vspace{-5mm}
\end{table*}

\subsection{Main Results}

Table \ref{tab:main_instruction} shows the performance of \method on five instruction-following benchmarks.

\paragraph{\method Outperforms All Previous Methods}
In the strong-to-weak setting, \method demonstrates performance that is comparable to or exceeds previous methods across all datasets. By fine-tuning on our generated data, \method achieves substantial improvements, particularly on IFEval and Multi-IF. When compared to strong baselines like AutoIF \citep{dong2024self}, \method achieves scores of 53.97 (Pr(S)) and 64.15 (Ins(S)) on IFEval and 81.91 (DRFR) on InfoBench, surpassing AutoIF by margins ranging from 1.29\% to 6.84\%. These results underscore \method's capability to effectively follow instructions, even with lower training data, representing a significant advancement over state-of-the-art approaches.

\paragraph{Iterative DPO Boosts Performance Effectively}
As shown in Table \ref{tab:main_instruction}, the iterative DPO process substantially enhances alignment with complex instructions. Specifically, in comparison to SFT, iterative DPO achieves an average improvement of 5\% in the strong-to-weak setting and 3.8\% in the self-alignment setting on MultiIF. Furthermore, this process enables \method to surpass state-of-the-art methods in three benchmarks that require LLM-based evaluation, with an improvement of 1.5\% on InfoBench, 4.6\% on LiveBench, and 2.62\% on FollowBench, demonstrating the importance of \method in handling diverse instructions.

\paragraph{Smaller Supervisor Yields Better Performance}
The self-alignment setting, which employs smaller model as supervisor, achieves superior performance relative to the strong-to-weak setting. This divergence is particularly evident during the SFT stage, wherein self-alignment outperforms strong-to-weak on IFEval, Multi-IF and LiveBench. While improvements introduced by DPO remain relatively incremetal, self-alignment still exhibits superior performance on two benchmarks. 
These results align with prior research by \citet{hui2024smaller,zhang2025best,li2025small}, which demonstrates that self-generated responses more closely to the distribution of the base model.

\paragraph{\method Achieves A New Milestone}
By scaling up the training data, \method achieves a new milestone in instruction-following alignment. With 181k data in the SFT stage and 20k data in the DPO stage, \method reaches impressive performance, with 71.35 (Pr(S)) and 79.38 (Ins(S)), while the LLaMA-3.1-8B-Instruct model only achieves 69.13 (Pr(S)) and 77.46 (Ins(S)), and comparable across the left benchmarks. This demonstrates that \method, when optimized and trained on larger datasets, not only improves instruction-following capabilities but also comes closest to matching the performance of LLaMA-3.1-8B-Instruct, marking a significant leap forward in model performance.

\begin{table*}[!ht]
    \centering
    \small
    \begin{tabular}{lccccc}
        \toprule
        \multirow{2}{*}{\textbf{Method}} & \textbf{Code} & \textbf{Reasoning} & \textbf{Math} & \textbf{Conversation} & \textbf{General }\\ 
        \cmidrule(r){2-2}  \cmidrule(r){3-3}  \cmidrule(r){4-4}  \cmidrule(r){5-6} \cmidrule(r){6-6}
        ~ & \textbf{HumanEval} & \textbf{BBH} & \textbf{GSM8k} & \textbf{Arena Hard}  & \textbf{LiveBench [All]} \\ 
        \midrule
        \rowcolor{green!5} LLaMA-3.1-8B-Instruct & 65.24  & 68.54  & 80.80  & 18.30  & 25.90 \\ 
        AutoIF \citeyearpar{dong2024self} & 46.34  & 67.18  & \textbf{51.50}  & 9.20 & 17.50  \\
        \midrule
        \method + SFT & 43.90 & 67.33 & 48.60 & 12.20 & 21.30  \\ 
        \hspace{15pt} + Iterative DPO & \textbf{47.56} &  \textbf{68.03} & 48.10 & \textbf{16.00} & \textbf{21.70} \\ 
        \hdashline[2pt/3pt]
        + SFT $scale\ up$  & 52.44  & 67.26  & 66.70  & 16.00   & 22.80  \\ 
        \rowcolor{blue!5}  \hspace{15pt} + Iterative DPO & \textbf{55.49}  & \textbf{68.44}  & \textbf{68.00}  & \textbf{31.40}  & \textbf{23.10}   \\ 
        \bottomrule
    \end{tabular}
    \caption{The general performance on mathematical, reasoning, coding, and conversational domains. We report Pass@1 on HumanEval, Acc on BBH and GSM8k, and Win Rate on Arena Hard.}
    \label{tab:main_cross}
    \vspace{-3mm}
\end{table*}

\begin{table*}[ht]
    \centering
    \small
    \resizebox{0.98\linewidth}{!}{
    \begin{tabular}{lcccccccc}
    \toprule
    \multirow{2}{*}{\textbf{Iteration}} &  \multicolumn{4}{c}{\textbf{IFEval}} & \multicolumn{3}{c}{\textbf{Multi-IF}} & \textbf{LiveBench} \\ 
    \cmidrule(r){2-5}  \cmidrule(r){6-8} \cmidrule(r){9-9} 
    & Pr(S) & Pr(L) & Ins(S) & Ins(L) & Turn1 & Turn2 & Turn3 & Score \\ 
    \midrule
    \textit{Iter 1} & 55.45$_{+1.48}$ & 61.55$_{+2.96}$ & 65.10$_{+0.95}$ & 70.74$_{+1.92}$ & 56.13$_{+3.58}$ & 32.11$_{+2.77}$ & 24.38$_{+2.09}$ & 42.20$_{+0.00}$ \\
    \textit{Iter 2} & 55.08$_{+1.11}$ & 62.66$_{+4.07}$ & 65.47$_{+1.32}$ & 71.82$_{+3.00}$ & 57.26$_{+4.71}$ & 34.92$_{+5.58}$ & 26.28$_{+4.00}$ & 47.20$_{+5.00}$ \\
    \textit{Iter 3} & 56.75$_{+2.78}$ & 63.03$_{+4.44}$ & 66.79$_{+2.64}$ & 72.42$_{+3.60}$ & 57.10$_{+4.55}$ & 34.87$_{+5.53}$ & 26.11$_{+3.82}$ & 45.70$_{+3.50}$ \\
    \textit{Iter 3$_{w. NCA}$}  & 58.22$_{+4.25}$ & 65.25$_{+6.66}$ & 68.11$_{+3.96}$ & 74.22$_{+5.40}$ & 58.14$_{+5.59}$ & 35.65$_{+6.31}$ & 26.55$_{+4.26}$ & 49.50$_{+7.30}$ \\
    \bottomrule
    \end{tabular}
    }
    \caption{The performance compared to the SFT model across each iteration during the Iterative DPO process.}
    \label{tab:iterative_dpo}
    \vspace{-3mm}
\end{table*}

\subsection{Cross-Domain Validation}
Table \ref{tab:main_cross} presents a comparative evaluation of \method across four general domains against AutoIF and LLaMA-3.1-8B-Instruct. Although \method exhibits slightly lower performance than AutoIF on math, it achieves substantial improvements on code and conversation. These gains are further amplified by scaling up the training data, and the application of the DPO stage consistently enhances performance across all evaluated domains.
In particular, \method contributes significantly to improving general capabilities, as evidenced by its performance on the comprehensive LiveBench benchmark \citep{white2024livebench} and the ArenaHard conversational benchmark \citep{li2024crowdsourced}. \method surpasses AutoIF by a statistically significant margin of 4.2\% on LiveBench and achieves a substantial 15.4\% improvement in conversational performance on ArenaHard. These results highlight the effectiveness of \method in advancing the development of more versatile and general models.

\section{Analysis}
\subsection{Impact of the Iterative DPO Process}
\begin{figure}[t]
    \centering
    \includegraphics[width=\linewidth]{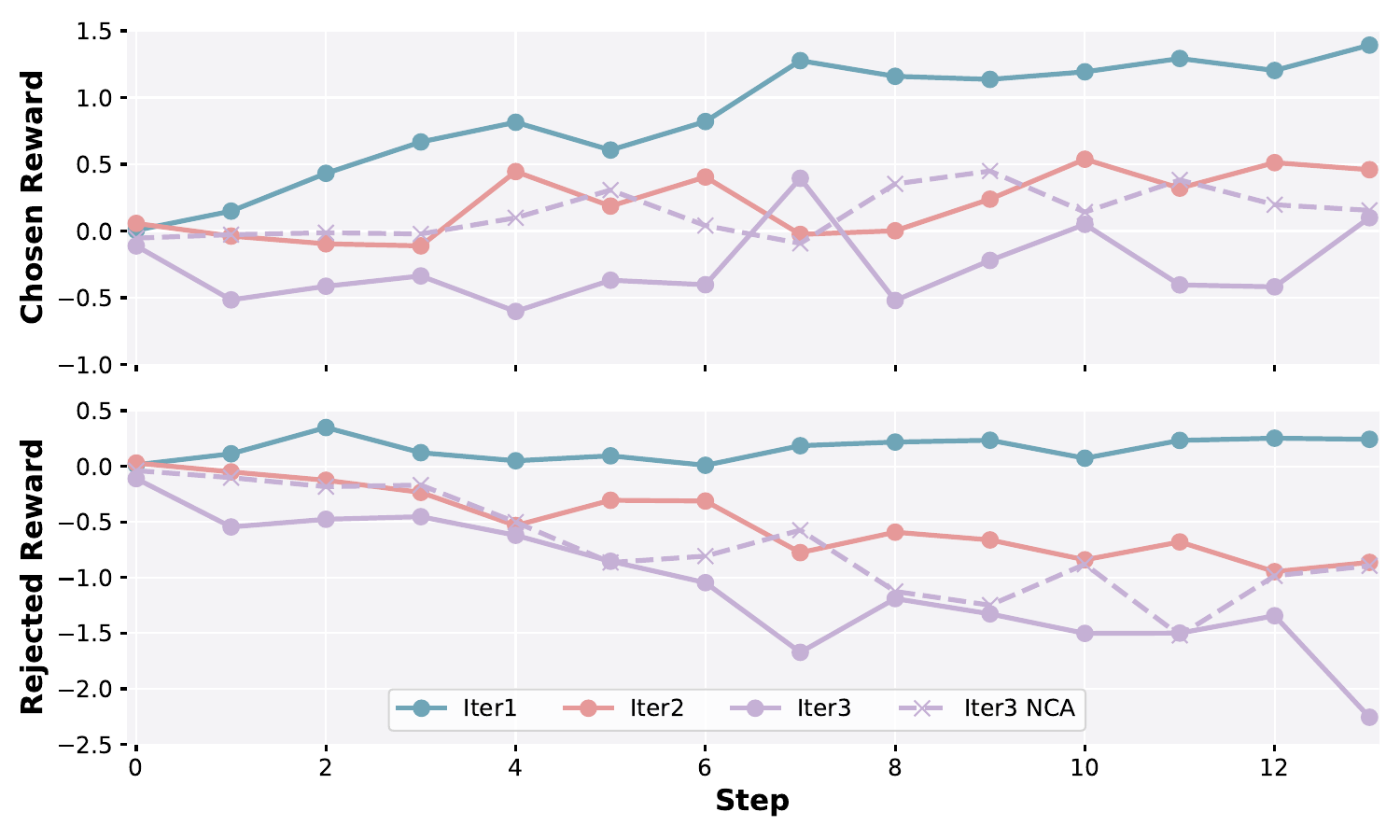}
    \caption{Reward trajectories for the chosen and rejected samples across the steps of each iteration during the Iterative DPO process.}
    \label{fig:dpo_loss}
    \vspace{-3mm}
\end{figure}

\noindent Given that \textit{UltraComposer} enables the iterative incorporation of constraints into instructions, we adopt an Iterative DPO training strategy in which instruction complexity is gradually increased over successive training stages. Figure~\ref{fig:dpo_loss} illustrates the reward trajectories for the chosen and rejected samples. As training progresses, the reward margin between chosen and rejected samples widens. However, by the third iteration, the optimization objective begins to diverge, prioritizing the maximization of the reward margin over improving the absolute reward values. This shift leads to both chosen and rejected sample rewards falling below zero, ultimately degrading model performance on three benchmarks, as shown in Table~\ref{tab:iterative_dpo}. To address this, we replace the DPO objective with NCA~\citep{chen2024noise}, which stabilizes the training dynamics and results in more consistent and robust performance across benchmarks.

\begin{table}[t]
    \centering
    \small
    \begin{tabular}{lccc}
        \toprule
        \textbf{Setting} & \textbf{c=1} & \textbf{c=2} & \textbf{c=3} \\
        \midrule
        \textbf{\textit{Strong-to-Weak}} &  91.76	& 86.41	& 79.44 \\
        \textbf{\textit{Self-Alignment}} & 92.46	& 89.57	& 85.79 \\
        \bottomrule
    \end{tabular}
    \caption{The pass rate of of \method across increasing instruction complexity levels during SFT stage. }
    \label{tab:sft_pass}
    \vspace{-3mm}
\end{table}

\subsection{Analysis of Sampling Efficiency}
We also evaluate the sampling efficiency of \method by measuring the proportion of generated responses that satisfy the filter question criteria, as instruction complexity increases through iterative applications of \textit{UltraComposer}. Table~\ref{tab:sft_pass} presents the overall pass rates of \method across varying constraint levels ($c=1,2,3$). The pass rate decreases with increasing instruction complexity. Notably, the self-alignment setting consistently outperforms the strong-to-weak setting. These trends are consistent with findings from prior studies \citep{hui2024smaller,zhang2025best,li2025small}.

Furthermore, Appendix~\ref{sec:scale_analysis} compares the data synthesis pass rates of \method and AutoIF, demonstrating the superior scalability and cost-efficiency of \method in dataset construction.
To assess the reliability of LLM-based evaluation during filtering, we conduct a human evaluation study, where we randomly sample 75 examples from the SFT dataset and assign each to five human annotators. The resulting agreement shows that LLM evaluations achieve an accuracy of approximately 80\%, which we consider sufficient for practical use.

\subsection{Contamination Analysis}
To ensure the integrity of our evaluation, we conduct a comprehensive contamination analysis of the training data generated by both AutoIF and \method across three benchmarks. We use a traditional n-gram overlap method ($overlap\ ratio=\frac{\# matched\_ngrams}{\# total\_ngrams}$) to identify any exact or near-duplicate sequences between training and test sets and utilize the LLM-based contamination detection framework from LM-Sys \citep{yang2023rethinking}, which leverages advanced LLMs to identify semantically rephrased versions of test instances in the training data.
Table~\ref{tab:contamination} reveals extremely low contamination rates across all configurations. 
Notably, both detection methods show that \method exhibits contamination levels comparable to those of AutoIF. These findings confirm that the self-generated training data is clean and does not compromise the validity of our evaluation results.

\begin{table}[t]
    \centering
    \small
    \resizebox{0.98\linewidth}{!}{
    \begin{tabular}{lcccc}
        \toprule
        \multirow{2}{*}{\textbf{Method}} & \multirow{2}{*}{\textbf{\#Data}}  & \textbf{IFEval}	& \textbf{Followbench} & \textbf{Multi-IF} \\
        \cmidrule(r){3-3} \cmidrule(r){4-4} \cmidrule(r){5-5}
        & & 541 & 944 & 4501 \\
        \midrule
        \multicolumn{5}{c}{\textit{\textbf{N-gram (N=13) Overlap Ratio}}} \\
        AutoIF & 10k  & 	0.0000 & 0.0000 & 0.0000 \\
        \method & 10k & 0.0000 & 0.0026 & 0.0000    \\
        \method & 181k & 0.0000 & 0.0033 & 0.0000    \\
        \midrule
        \multicolumn{5}{c}{ \textit{\textbf{LLM Decontaminator (Rephr.)}}} \\
        AutoIF & 10k & 0.0000 & 0.0032 & 0.0000 \\
        \method & 10k & 0.0018 & 0.0117 & 0.0000 \\
        \method & 181k & 0.0166 & 0.0360 & 0.0082 \\ 
        \bottomrule
    \end{tabular}
    }
    \caption{ Contamination analysis on SFT data generated by AutoIF and \method on three benchmarks.}
    \label{tab:contamination}
    \vspace{-3mm}
\end{table}

\begin{figure}[t]
    \centering
    \includegraphics[width=0.9\linewidth]{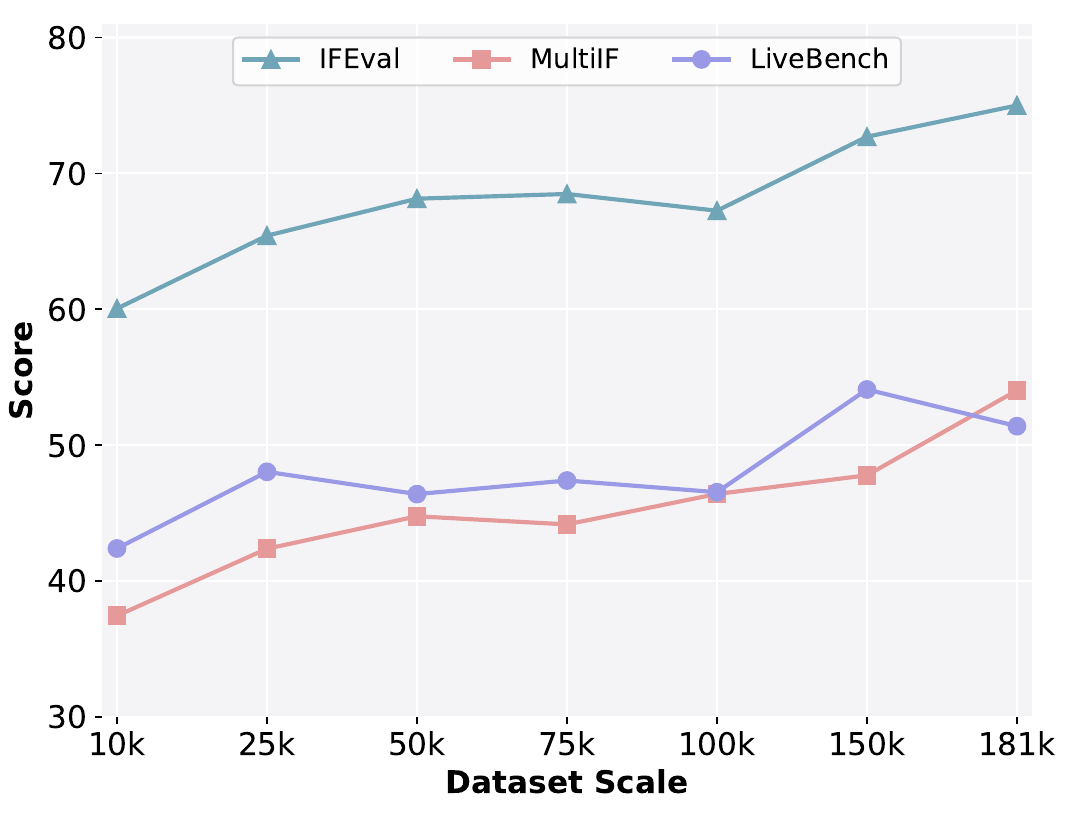}
    \caption{Scaling the training set on SFT stage.}
    \label{fig:scale}
    \vspace{-5mm}
\end{figure}

\subsection{Scalability of \method}
To validate the effieiency and effectiveness of \method, we conduct scaling up experiments under the self-alignment setting. Figure~\ref{fig:scale} shows the impact of varying training data sizes during the SFT stage. With about 181k training samples, \method demonstrates powerful performance compared to baselines, highlighting its strong capacity to scale with increasing data volume. Moreover, we analyze the impact of multi-turn data in Appendix \ref{sec:multi_turn}, and incoporate such data in our scaling experiments.

\subsection{Ablation Studies on \method}
\begin{figure}[t]
    \centering
    \includegraphics[width= \linewidth]{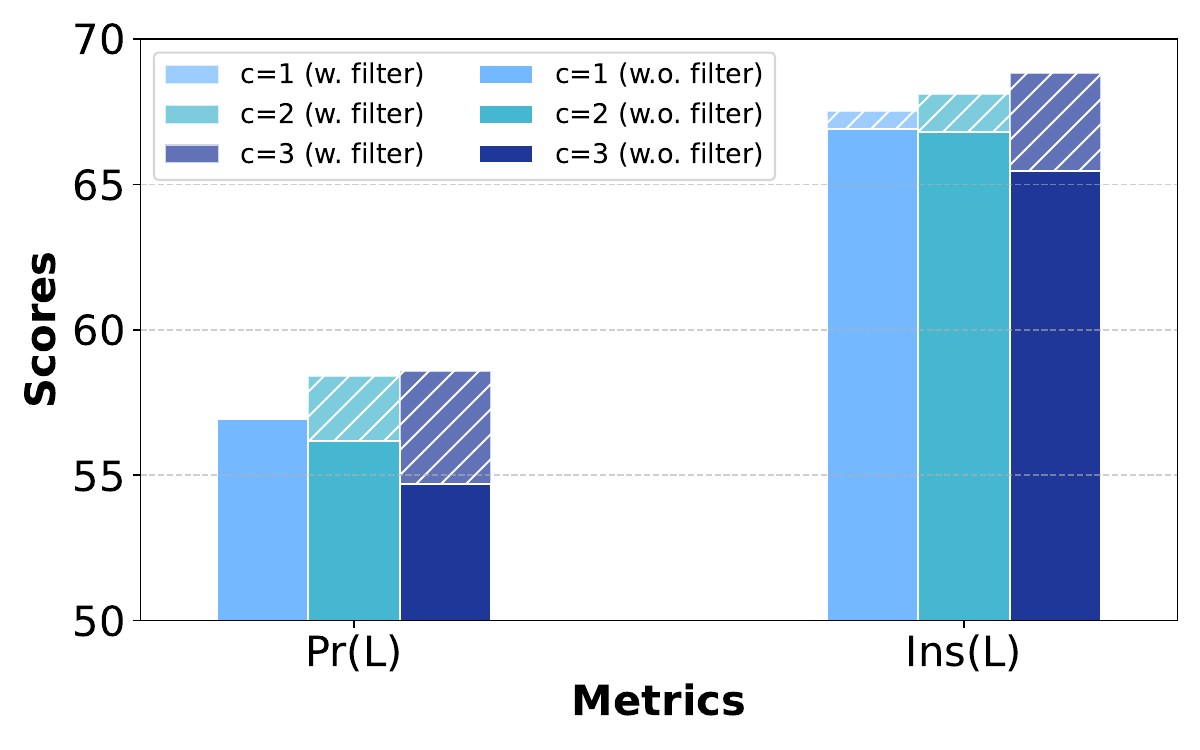}
    \caption{Ablations on the number of added constraints and the evaluation question filter.}
    \label{fig:ablation_constraint}
    \vspace{-5mm}
\end{figure}

The iterative augmentation capability of \textit{\composer} raises a critical question for the SFT stage: should simple or complex instructions be prioritized for training? Figure \ref{fig:ablation_constraint} presents the results of using varying levels of instruction complexity during the SFT stage. The results demonstrate that as instruction complexity increases, performance correspondingly improves, reaching the peak after three iterations with.
Furthermore, we evaluate the effectiveness of our evaluation questions. Without filtering out low-quality responses, performance deteriorates significantly over 3.35\%-5.36\%. This mechanism becomes increasingly critical as instruction complexity grows, with the performance gap widening alongside the increasing complexity, underscoring the importance of this module in maintaining high-quality training data. We provide cases to illustrate the augmented instructions and evaluation questions in Appendix \ref{sec:case_study}.

\subsection{Extension of Self Alignment}
\begin{figure}[t]
    \centering
    \includegraphics[width=\linewidth]{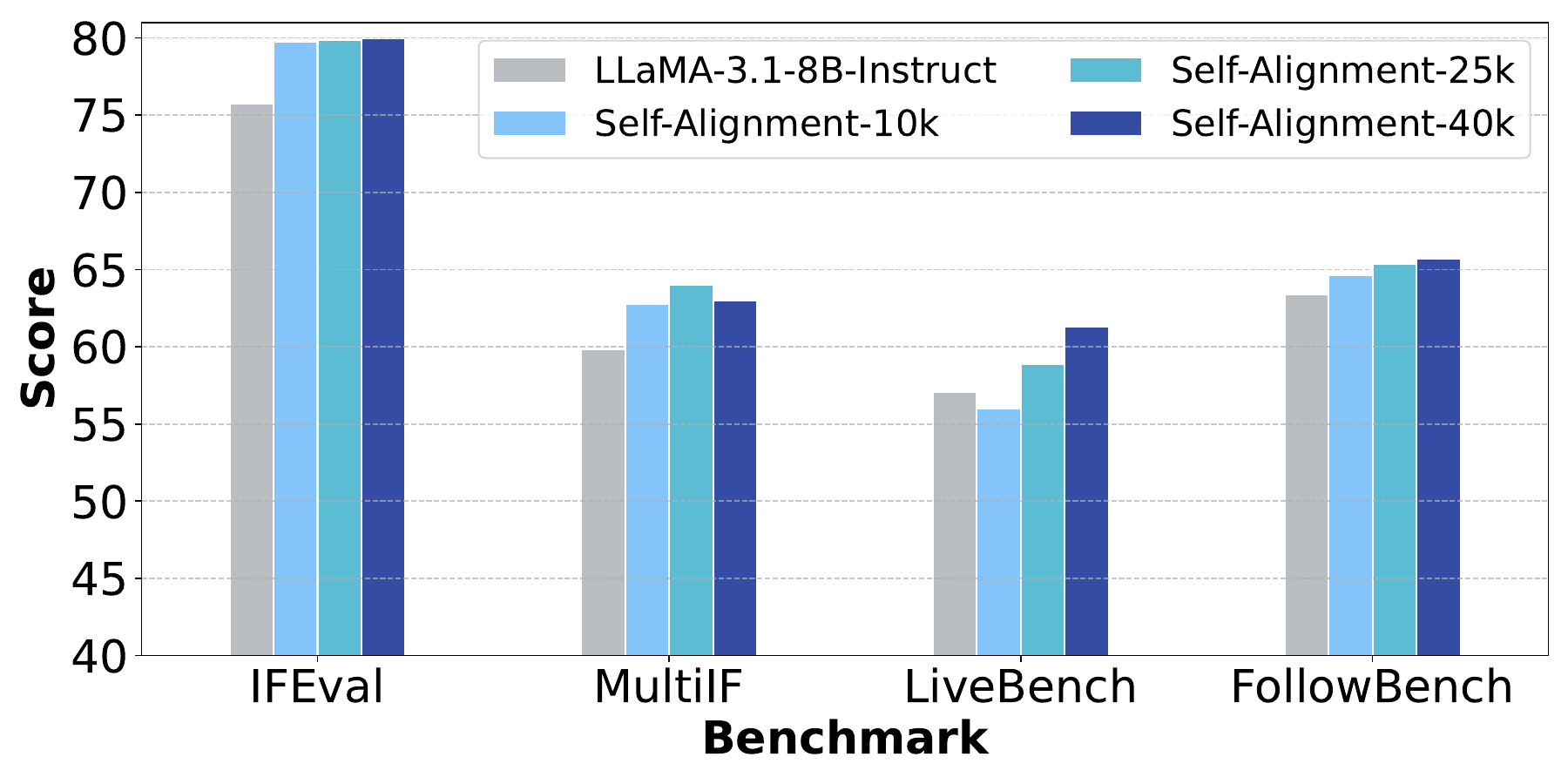}
    \caption{The performance of exploring the potentiality of \method on self-alignment.}
    \label{fig:self_evolve}
    \vspace{-4mm}
\end{figure}

In our main experiments, we distill knowledge from the Instruct version model to enhance the vanilla model, demonstrating the effectiveness of \method. However, the potential for \method to independently enhance a strong model like \citet{cheng2024spar} has not yet been explored. In this section, we conduct experiments to investigate the self-improvement capabilities of \method. Under the self-alignment setting, we use data generated by LLaMA-3.1-8B-Instruct to enable the model to train itself. As shown in Figure \ref{fig:self_evolve}, \method significantly boosts the performance of the strong model across different size of training data, even without a more powerful supervisor. Specifically, \method improves performance on IFEval by 2.4\%-5.9\%, on Multi-IF by 3.74\%-5.38\%, further validating the effectiveness of our approach.

\subsection{Generalizability of \method}

To assess the generalizability of \method across different foundation models, we apply it to the Qwen2-7B base model \citep{yang2024qwen2technicalreport}. As shown in Table~\ref{tab:qwen_instruction}, \method maintains strong performance when built on the Qwen2 architecture, demonstrating its adaptability to different model backbones. Notably, compared to AutoIF, \method exhibits greater potential in aligning LLMs with instruction-following capabilites.

\begin{table*}[!t]
    \centering
    \resizebox{\textwidth}{!}{
    \begin{tabular}{lc*{7}{>{\centering\arraybackslash}p{1cm}}*{3}{>{\centering\arraybackslash}p{1.7cm}}}
    \toprule
    \multirow{2}{*}{\textbf{Method}} & \multirow{2}{*}{\textbf{\#Data}} &  \multicolumn{4}{c}{\textbf{IFEval}} & \multicolumn{3}{c}{\textbf{Multi-IF}} & \textbf{LiveBench} & \textbf{FollowBench} \\ 
    \cmidrule(r){3-6}  \cmidrule(r){7-9} \cmidrule(r){10-10}  \cmidrule(r){11-11} 
    & & Pr(S) & Pr(L) & Ins(S) & Ins(L) & Turn1 & Turn2 & Turn3 & Score & SSR \\ 
    \midrule
    \rowcolor{green!5}Qwen2-7B-Instruct & - & 52.68 & 55.63   & 62.82  & 65.34   & 54.44  & 39.41  & 29.95  & 46.30  & 63.36\\
    AutoIF\citeyearpar{dong2024self}$^\dag$ & 10k & 40.70 & 44.50 & 51.30 & 55.40 & - & - & - & - & 53.30\\ 
    \midrule
    \method + SFT & 10k & 44.17   & 47.31   & 54.19    & 57.55    & 47.56    & 25.38   & 18.13    & 35.20  & 54.16 \\
    \hspace{6pt} + Iterative DPO & 8k & 45.28   & 48.61    & 56.59   & 59.23    & 48.57   & 28.45   & 19.60   & 39.80  & 55.44\\
    \midrule
    \end{tabular}
    }
    \caption{The results on four instruction-following benchmarks with Qwen2-7B as the backbone model.
    }
    \label{tab:qwen_instruction}
    \vspace{-4mm}
\end{table*}

%% file: sections/relatedwork.tex
\section{Related Work}
\subsection{Instruction Following}
Instruction following is a core area for LLMs, aiming to improve understanding and execution of complex human instructions. Early work \citep{wei2021finetuned,no_robots,jiang2023followbench} use curated datasets of human-written instructions and responses. Recent methods automate this using LLMs, \citet{xu2023wizardlm} and \citet{sun2024conifer} prompt LLMs to evolve or complicate instructions. However, this can yield low-quality data due to LLMs' limitations. To improve quality, \citet{wang-etal-2023-self-instruct,dong2024self} add human priors like verifiable constraints, but this reduces instruction diversity.
In contrast, \method decomposes user instructions into constraints and evaluation questions, then trains \textit{\composer} to generate diverse, complex instructions with accurate responses, offering a robust approach to instruction data generation.

\subsection{Perference Learning}

Preference learning has emerged as a key method to improve instruction-following by refining models through feedback \citep{NEURIPS2022_b1efde53, dong2024self, sun2024conifer, gao2024unifiedviewpreferencelearning,si-etal-2025-aligning}. It typically enhances models finetuned on instruction data using reward signals from human or automated to guide learning. While RLHF with PPO is common, it depends on ranked responses, which are costly and labor-intensive. Recent work \citep{rafailov2024direct, chen2024noise} addresses this via direct preference optimization, reducing reliance on human input.
\method supports this by generating evaluation questions that guide preference learning more efficiently. It complements direct optimization with a scalable, cost-effective approach to producing instruction-following data.

%% file: sections/appendix.tex
\section{Experimental Setup}
\subsection{Datasets and Baselines}
\label{sec:dataset}
\subsubsection{Datasets}

\noindent \textbf{ShareGPT} \footnote{\url{https://huggingface.co/datasets/shibing624/sharegpt\_gpt4}} is an open-source and multi-turn conversation dataset that contains over 52K user-shared chatting histories with GPT-4. We decompose the human instructions from ShareGPT into around 200K data pairs to train our UltraComposer. For main experiment, we randomly select 10K human instructions to generate augmented instructions.

\noindent  \textbf{OpenHermes} \citep{OpenHermes} is a large-scale, diverse, and high-quality compilation consisting of around 1M synthetically generated instruction and chat samples. We randomly select a subset of 150K instructions from OpenHermes-2.5 for scaling experiment.

\noindent  \textbf{No Robots} \citep{no_robots} is a high-quality dataset of 10k instructions and demonstrations created by skilled human annotators. We use all instructions from No Robots for scaling experiment.

\subsubsection{Baselines}
\noindent \textbf{AutoIF} \citep{dong2024self}  uniquely employs code verification to conducting scalable and reliable data generation for complex instruction-following in LLMs. We reproduced AutoIF utilizing the official open-source code\footnote{\url{https://github.com/QwenLM/AutoIF}}.

\noindent \textbf{Conifer} \citep{sun2024conifer} curates a novel instruction tuning dataset which aims to enhance how LLMs, particularly open-source models, follow complex instructions involving multiple, intricate constraints. We generate responses for the public data using LLaMA-3.1-70B-Instruct\footnote{\url{https://huggingface.co/datasets/ConiferLM/Conifer}}.

\noindent \textbf{Evol-Instruct} \citep{xu2023wizardlm} automatically mass-produces high-complexity training data by generating diverse instructions with varying difficulty levels. We sample 10k data and generate responses\footnote{\url{https://huggingface.co/datasets/WizardLMTeam/WizardLM_evol_instruct_70k}}.

\noindent \textbf{SPaR} \citep{cheng2024spar} proposes a self-play framework to enhance instruction-following abilities for LLMs, where an LLM refines its own responses via tree-search. We reimplement SPaR utilizing its official open-source dataset\footnote{\url{https://huggingface.co/datasets/CCCCCC/SPaR}}.

\subsection{Evaluation Benchmarks}
\label{sec:benchmark}
\noindent \textbf{IFEval} \citep{zhou2023instruction} is an easy-to-produce benchmark designed to evaluate the instruction-following capability of LLMs. IFEval constructs around 500 prompts that contain 25 types of verifiable instructions. We use both loose and strict accuracy metrics at prompt and instruction levels in our evaluation. 

\noindent \textbf{Multi-IF} \citep{he2024multi} is a benchmark designed to assess LLMs' proficiency in following multi-turn and multilingual instructions. Based on IFEval, Multi-IF contains 4,501 multilingual conversations, where each has three turns. We report the average accuracy across all languages for each of the three rounds in the experiment.

\noindent \textbf{InfoBench} \citep{qin2024infobench} is a benchmark comprising 500 diverse instructions and 2,250 decomposed questions across multiple constraint categories, adopting a new
metric Decomposed Requirements Following Ratio (DRFR) for evaluating LLM's ability to follow instructions. We use GPT-4-1106-preview as the evaluator in our assessments.

\noindent \textbf{FollowBench} \citep{jiang2023followbench} is a multi-level fine-grained constraints following benchmark for LLMs.  FollowBench incorporates five distinct fine-grained constraint types (Content, Situation, Style, Format, and Example) and underscores multi-level mechanism when building instruction prompts. In our experiment, we prompt the GPT-4-1106-preview to assess whether LLM's outputs have satisfied each individual constraint.

\noindent \textbf{LiveBench} \citep{white2024livebench} is a LLM benchmark that contains a wide variety of challenging tasks (math, coding, reasoning, language, instruction-following, and data analysis) and automatically scores answers according to objective ground-truth values. When assessing the instruction-following skills of LLMs, we employ the instruction-following subset, while the entire dataset is utilized to gauge their overall capabilities.

\noindent \textbf{GSM8K} \citep{cobbe2021training} comprises 8.5K high-quality, multilingual grade school math word problems, specifically designed to assess the multi-step mathematical reasoning proficiency of language models. We report the overall accuracy in the experiment.

\noindent \textbf{HumanEval} \citep{chen2021evaluating} consists of 164 programming problems with function signatures, docstrings, bodies, and unit tests, averaging 7.7 tests per problem. It is utilized to evaluate the coding abilities of LLMs. HumanEval assesses the capability of LLMs in program synthesis from docstrings, testing language comprehension, reasoning, algorithms, and elementary mathematics skills. We report Pass@1 on HumanEval in the experiment.

\noindent \textbf{BBH} \citep{suzgun2022challenging} is a clean, challenging and tractable subset benchmark filtered from Big Bench, which includes 23 types of difficult tasks and 6,511 evaluation examples in total. BBH primarily assesses the models' reasoning capacities and problem-solving skills comprehensively. In the experiment we report the accuracy metrics.

\noindent \textbf{Arena Hard} \citep{li2024crowdsourced} is an automatic  LLM benchmark consisting of 500 challenging challenging user queries, which is curated to evaluate the comprehensive performance of LLMs in user dialogue scenarios. In the experiment, we adopt GPT-4-1106-preview as the judge model and report the win rate of our models against the baseline model (GPT-4-0314).

\subsection{Training Strategies}
\label{sec:strategy}

\method offers flexible training strategies for aligning model with instruction following capabilities. To thoroughly evaluate the effectiveness, we provide two approaches:

\paragraph{Supervised Finetuing (SFT).}
Given the dataset $\mathcal{D}_{data}$, we apply standard Supervised Finetuning (SFT) objective on vanilla model $\pi$ with parameters $\theta$, as shown in Eq. \ref{eq:4}: 
\begin{equation}
\label{eq:4}
    \mathcal{L}_{SFT}(\pi_\theta)=\sum_{(\bar{x}, y_{c})\in \mathcal{D}_{data}} \log\ \pi_{\theta}(y_{c}|\bar{x})
\end{equation}
where $\bar{I}$ represents the augmented instruction, and $r_{c}$ denotes the corresponding chosen response.

\paragraph{SFT + Iterative Online DPO.} As \method is equipped with evaluation questions, it facilitates quality control by enabling the generation of pairwise responses with varying quality levels. This property makes it particularly suitable for the application of Direct Perference Optimization (DPO, \citet{rafailov2024direct}) to refine the fine-tuned model, $\pi_{ref}$. The DPO objective is formulated as Eq. \ref{eq:5}:
\begin{equation}
\label{eq:5}
\begin{aligned}
    \mathcal{L}_{DPO}(\pi_\theta, \pi_{ref}) = - \mathbb{E}_{(\bar{x}, y_{c}, y_{r}) \in \mathcal{D}_{data}} \log\ \sigma(\beta\ \cdot \Delta)\\
    \Delta = (\log \frac{\pi_\theta(y_c|\bar{x})}{\pi_{ref}(y_c|\bar{x})} - \log\frac{\pi_\theta(y_r|\bar{x})}{\pi_{ref}(y_r|\bar{x})})
\end{aligned}
\end{equation}
where $\beta$ is a scaling hyperparameter, $\sigma$ denotes the sigmoid function, and $\pi_\theta$ is initialized from $\pi_{\text{ref}}$ and further optimized during the DPO stage.

In the context of \method, the \textit{\composer} enables an iterative augmentation of instructions, transitioning from simpler to more complex tasks. This allows the DPO process to be formulated as an iterative curriculum. At each iteration, the model $\pi_{\text{ref}}$ is replaced with the latest optimized model from the previous stage. Concurrently, more challenging instruction-following datasets are generated and utilized for further training. This iterative approach ensures continuous improvement in model performance and adaptability across increasingly complex scenarios.

Moreover, during the iterative process, as observed by \citep{chen2024noise}, the DPO objective primarily focuses on optimizing the margin between the chosen and rejected samples, rather than directly maximizing the probability of chosen samples and minimizing that of the rejected ones. To address this, we employ the Noise Contrastive Estimation (NCA, \citet{chen2024noise}) loss in the final iteration, and the objective is defined in Eq. \ref{eq:6}:
\begin{equation}
\label{eq:6}
\begin{aligned}
    \mathcal{L}_{NCA}(\pi_\theta, \pi_{ref}) &= \\
    &\hspace{-2cm} - \mathbb{E}_{(\bar{x}, y_{c}, y_{r}) \in \mathcal{D}_{data}}  \bigg[ 
    \log\ \sigma (\beta\log \frac{\pi_\theta(y_c|\bar{x})}{\pi_{ref}(y_c|\bar{x})} ) \\
    &\hspace{-2cm}  + \frac{1}{2} \sum_{y \in \{y_c, y_r\}} 
    \log\ \sigma (-\beta\log \frac{\pi_\theta(y|\bar{x})}{\pi_{ref}(y|\bar{x})} ) \bigg]
\end{aligned}
\end{equation}

\subsection{Implementation Details}
\label{sec:train_details}

Our experiments are conducted on 8$\times$A100 GPUs (80GB) using mixed precision with bf16, DeepSpeed ZeRO Stage 3 \citep{rasley2020deepspeed}, and FlashAttention 2 \citep{dao2023flashattention}.

In the SFT stage, we perform full fine-tuning with a learning rate of 1e-5. The maximum token length is set to 8192 and variable-length packing is enabled. We use AdamW \citep{loshchilov2017decoupled} as the optimizer with a warmup ratio of 0.03 and employ a LinearLR scheduler at the beginning, transitioning to CosineAnnealingLR towards the end.

In the DPO stage, the configuration is similar, with the only difference being a lower learning rate of 5e-7. Additionally, the beta parameter of DPO loss is set to 0.1.

\section{Prompts of \method}
\label{sec:prompt}
To train our \textbf{\textit{\composer}}, we prompt LLM to perform \textbf{Instruction Decomposition} and \textbf{Eval Question Generation}.

We use the following prompt template to decompose human instructions:

\begin{tcolorbox}[title = {Prompt Template of Instruction Decomposition}, breakable]
You are an expert in extracting instruction constraints from a given query. \\
\textbf{Definition of Constraint:} The smallest unit of restriction or requirement that the instruction imposes on the task.

\textbf{Query:} \{query\}

\begin{itemize}
    \item If the query is not a question, or is simple or straightforward without any constraints, please only respond with the following JSON, indicating that no constraints are present.
    \begin{verbatim}
    {
        "Complex": False
    }
    \end{verbatim}
    \item If constraints are present, follow these steps:
    \begin{enumerate}
        \item Identify the Basic Query: Clearly understand the primary goal of the query, stripping away any constraints. The Basic Query should be the essential task without any added conditions or restrictions.
        \item Extract and Categorize Constraints: Identify and classify constraints based on the following types:
        \begin{itemize}
            \item Content Constraints:
            \begin{itemize}
                \item Specific Terms or Symbols: Mandatory use of certain terms or symbols with their exact placement (e.g., must include the word 'beautiful').
                \item Required Elements or Concepts: Mandates for including specific elements or concepts in responses, reflecting a scenario or object (e.g., highlights the Great Wall).
                \item Thematic Directives: Instructions related to thematic content, perspective, or tone, emphasizing response significance (e.g., write a poem about London).
            \end{itemize}
            \item Numerical Constraints:
            \begin{itemize}
                \item Constraints on quantities related to the content, such as the number of points, sentences, paragraphs, response length, or examples (e.g., within a single paragraph with three sentences).
            \end{itemize}
            \item Stylistic Constraints:
            \begin{itemize}
                \item Desired tone and style for the response (e.g., formal, informal, conversational).
                \item Specific language or terminology to be used or avoided (e.g., encyclopedic style).
            \end{itemize}
            \item Format Constraints:
            \begin{itemize}
                \item Required structure or format for the response (e.g., list, JSON, bullet points, Java language).
                \item Presentation styles or formatting requirements (e.g., electronic medical record format).
            \end{itemize}
            \item Linguistic Constraints:
            \begin{itemize}
                \item Language use in specific contexts, such as discourse, dialects, sociolects, and language policies (e.g., in English).
                \item Sentence structure, including phrases, constituents, and the use of imperatives (e.g., with nouns and verbs).
                \item Internal structure of words, including roots, affixes, and morphological changes (e.g., lowercase, single-rhyme).
            \end{itemize}
        \end{itemize}
        \item Response Format:
        \begin{itemize}
            \item Do not consider details that are part of the content itself, such as those used in descriptions, scenarios, or examples, unless they directly impose a restriction of the response.
            \item The Basic Query should represent the query’s core goal, free from any constraints. 
            \item Ensure that extracted constraints do not overlap with the Basic Query.
            \item Present each constraint as a dictionary within a list, where each dictionary contains:
            \begin{itemize}
                \item \texttt{'constraint'}: The specific restriction or requirement.
                \item \texttt{'simplified query'}: The query after removing this constraint, polished for coherence and correctness.
            \end{itemize}
            \item Exclude any constraint types not present in the query.
        \end{itemize}
    \end{enumerate}
    \end{itemize}

\begin{verbatim}
{
    "Complex": True,
    "Basic Query": ...,
    "Content Constraints": [
        {
            "constraint": "...",
            "simplified query": "..."
        },
        {
            "constraint": "...",
            "simplified query": "..."
        },
    ],
    ...
}
\end{verbatim}
Please only provide the response in JSON format.
\end{tcolorbox}

We use the following prompt template to generate evaluation questions for instructions:

\begin{tcolorbox}[title = {Prompt Template of Eval Question Generation}, breakable]
You are an expert in crafting questions to evaluate whether a response to a query adheres to specific constraints.

For the given constraint, please design a question that human evaluators can use to assess if the response meets the specified constraint. The question should focus solely on the given constraint and not other constraints present in the original query.

Specifically, if the given constraint is meaningless or is a part of the content itself, such as those used in descriptions, scenarios, or examples, you can respond with an empty string.

\textbf{Query:} \{query\} \\
\textbf{Constraint:} \{constraint\}

Please design a question for the specified constraint for the given query, and respond in the JSON format without explanation.

\begin{verbatim}
{
    "question": "string",
}
\end{verbatim}
\end{tcolorbox}

We use the following template to train \textit{\method}

\begin{tcolorbox}[title = {Prompt Template of UltraComposer},breakable]
Input:

[history]: ...

[initial query]: ...

Output:

\begin{verbatim}
{
    "augmented query": .., 
    "question": ...
}
\end{verbatim}
\end{tcolorbox}

For \textbf{Generate-then-Evaluate} process, we prompt LLM to perform \textbf{Response Generation} and \textbf{Response Evaluation}.

First we use the following prompt template to generate responses for the augmented instructions:
\begin{tcolorbox}[title = {Prompt Template of Response Generation},breakable]
You are an expert tasked with answering the given query. Please provide a clear and concise response directly, without introductory phrases such as 'What a great question,' 'Here is the answer,' or similar expressions. Focus solely on addressing the query.

Now please answer the given query while strictly following its inside constraints.

\textbf{[Query]} \{query\}
\end{tcolorbox}

Then we use the following prompt template to evaluate the quality of those generated responses:
\begin{tcolorbox}[title = {Prompt Template of Response Evaluation}, breakable]
You are an expert that is good at judging whether the response to a given query meets the specified evaluator questions. \\
Your task is to carefully examine the response to determine if it adheres to each requirement outlined in the evaluator questions.

\textbf{[Query]} \{query\} \\
\textbf{[Response]} \{response\} \\
\textbf{[Evaluator Question]} \{question\}

For each question, please provide a justification for your evaluation, explaining how the response does or does not satisfy the criteria and a score (\texttt{'YES'} or \texttt{'NO'}) indicating whether the answer satisfies each constraint.

You should only respond in the following JSON format:
\begin{verbatim}
{
    "Question 1": {
        "explanation": "",
        "score": "YES" or "NO"
    },
    "Question 2": {
        "explanation": "",
        "score": "YES" or "NO"
    },
}
\end{verbatim}
\end{tcolorbox}

\section{Additional Experimental Results}

\subsection{Analysis of Sampling Efficiency}
\label{sec:scale_analysis}

Moreover, Table~\ref{tab:cost} further compare the pass rates during dataset synthesis, where \method demonstrates substantial improvements over AutoIF. Specifically, \method achieves an SFT pass rate of 85\% and a DPO pass rate of 60\%, compared to only 20\% and 26\%, respectively, for AutoIF. This indicates that for generating an equivalent amount of data, \method reduces costs by a factor of three to five. Furthermore, during the rejection sampling stage, while AutoIF necessitates rigorous function-based filtering for instruction synthesis and response generation, \method achieves this with a single LLM call, making it far more scalable and cost-efficient.

\begin{table}[h]
    \centering
    \small
    \begin{tabular}{lcc}
        \toprule
         \textbf{Method} & \textbf{SFT Pass Rate} & \textbf{DPO Pass Rate}  \\
         \midrule
         AutoIF & 20\%  & 26\% \\
         \method & 85\% & 60\% \\
         \bottomrule
    \end{tabular}
    \caption{The overall pass rate of data synthesis.}
    \label{tab:cost}
    \vspace{-5mm}
\end{table}

\subsection{Analysis of Multi-Turn Data}
\label{sec:multi_turn}
Building on prior work that emphasizes enhancing multi-turn instruction-following capabilities \citep{sun-etal-2024-parrot, he2024multi}, our analysis reveals that incorporating multi-turn data during the SFT stage significantly improves \method's performance across various benchmarks. As shown in Table \ref{tab:multiturn}, the inclusion of multi-turn data results in performance gains of 3.01\% on Multi-IF, 1.18\% on InfoBench, and 5.10\% on LiveBench, compared to the baseline SFT model without such data. These improvements highlight the critical role of multi-turn interactions in training, allowing the model to better understand conversational context and dependencies, thereby enhancing its instruction-following capabilities. Therefore, we incorporate multi-turn data in our scaling experiments.

\begin{table}[h]
    \centering
    \small
    \begin{tabular}{lccccc}
    \toprule
    \textbf{Method} & \textbf{Multi-IF} & \textbf{InfoBench} & \textbf{LiveBench} \\
    \midrule
    \method + SFT & 40.12 & 77.78 & 46.60 \\ 
    \hspace{3pt} $w.\ multi\ turn$ & 43.13 & 79.86 & 54.20 \\
    \hdashline[2pt/3pt]
    \rowcolor{blue!5} \hspace{20pt} $\Delta$ & +3.01 & +1.18 & +5.10 \\
    \bottomrule
    \end{tabular}
    \caption{The performance comparison of incorporating multi-turn data during the SFT stage.}
    \label{tab:multiturn}
    \vspace{-5mm}
\end{table}

\subsection{Case Study}
\label{sec:case_study}
By modeling the distribution of real-world instructions, \method supports effective instruction augmentation while minimizing inconsistencies between newly added constraints and the original instructions. 
Thus, \method eliminates the need to verify whether the constraints are consistent with the original instructions \citep{dong2024self,katz2024evolutionary}. Additionally, the evaluation questions take over a separate score-filtering stage. Consequently, \method achieves greater efficiency and incurs minimal costs when constraining large-scale datasets.

Table \ref{tab:t6} shows some examples of augmented instructions and evaluation questions generated by \method. The original queries come from ShareGPT dataset.

\begin{table*}
\begin{tabular}{p{5cm}p{5cm}p{5cm}}
\hline
\textbf{Original Query} & \textbf{Augmented Instruction} & \textbf{Eval Question} \\
\hline 
 Explain merkle tree in blockchain. & Explain merkle tree in blockchain \textbf{to a 10 years old}. & Is the explanation of a Merkle tree in the context of blockchain presented in a way that a 10-year-old can understand? \\
\hline
We are driving in a car. It is cold outside, windshield is frozen, and we are on a secluded road to a small village. This is scene from a horror movie, and it's just starting. Describe a scene in great detail. & We are driving in a car. It is cold outside, windshield is frozen, and we are on a secluded road to a small village. This is scene from a horror movie, and it's just starting. Describe a scene in great detail, and \textbf{write it in the style of a gothic horror author}. & Does the response evoke a dark, eerie, and ominous atmosphere, characteristic of gothic horror? \\
\hline
 Design a html form with form tags. & Design a html form with form tags \textbf{for the following 3 user inputs: first\_name, last\_name, date\_of\_birth}. & Does the HTML form include form tags for exactly three user inputs: first\_name, last\_name, and date\_of\_birth? \\
\hline
 I'm planning to visit Okinawa Japan from April 7th to April 10th. Do you have any recommendation on what to do while I'm there?& I'm planning to visit Okinawa Japan from April 7th to April 10th. Do you have any recommendation on what to do while I'm there? \textbf{I'd like to focus on nature, food, and local culture.}&Does the response recommend activities in Okinawa that focus on nature, food, and local culture? \\
\hline
 What is the meaning of life?&What is the meaning of life? \textbf{Explain it in 5 paragraphs}. &Is the response to the question explained in exactly 5 paragraphs? \\
\hline
Write a homepage for translation business.& Write me a homepage for translation business \textbf{in wordpress}.&Is the homepage for the translation business designed using WordPress? \\
\hline
\end{tabular}
\caption{Examples of \method's data pair.}
\label{tab:t6}
\end{table*}